\documentclass[sigconf]{acmart}

\def\BibTeX{{\rm B\kern-.05em{\sc i\kern-.025em b}\kern-.08emT\kern-.1667em\lower.7ex\hbox{E}\kern-.125emX}}
    
\copyrightyear{2019} 
\acmYear{2019} 
\setcopyright{acmlicensed}
\acmConference[AIES '19]{AAAI/ACM Conference on AI, Ethics, and Society}{January 27--28, 2019}{Honolulu, HI, USA}
\acmBooktitle{AAAI/ACM Conference on AI, Ethics, and Society (AIES '19), January 27--28, 2019, Honolulu, HI, USA}
\acmPrice{15.00}
\acmDOI{10.1145/3306618.3314273}
\acmISBN{978-1-4503-6324-2/19/01}

\newcommand{\ignore}[1]{}

\newcommand{\ted}{TED }

\begin{document}

%
\title{TED: Teaching AI to Explain its Decisions}

%
\author{
Michael Hind, 
Dennis Wei,
Murray Campbell, 
Noel C.\ F.\ Codella,
Amit Dhurandhar,
}
\author{
Aleksandra Mojsilovi\'c, 
Karthikeyan Natesan Ramamurthy,
Kush R.\ Varshney
}
\email{{hindm,dwei,mcam,nccodell,adhuran,aleksand,knatesa,krvarshn}@us.ibm.com}
%
\affiliation{%
  \institution{IBM Research AI}
  \streetaddress{1101 Kitchawan Road}
  \city{Yorktown Heights}
  \state{New York}
  \postcode{10592}
}

%

\renewcommand{\shortauthors}{Hind, Wei, Campbell, Codella, Dhurandhar, Mojsilovi\'c, et al.}


%
\begin{abstract}
Artificial intelligence systems are being increasingly deployed due to their potential to increase the efficiency, scale, consistency, fairness, and accuracy of decisions.  However, as
many of these systems are opaque in their operation, there is a
growing demand for such systems to provide explanations for their
decisions.  Conventional approaches to this problem attempt to expose
or discover the inner workings of a machine learning model with the
hope that the resulting explanations will be meaningful to the consumer.
In contrast, this paper suggests a new approach to this problem. It
introduces a simple, practical framework, called Teaching Explanations for Decisions (\emph{TED}), that
provides meaningful explanations that match the mental model of the
consumer.  We illustrate the generality and effectiveness of this
approach with two different examples, resulting in highly accurate explanations with no loss of prediction accuracy for these two examples.
\end{abstract}

%
%

%
\keywords{Explainable AI; Meaningful Explanation; AI Ethics;
Elicitation; Supervised Classification; Machine Learning}

%
\maketitle

\section{Introduction}\label{sec-intro}

Machine learning based systems have proven to be quite effective for
producing highly accurate results in several domains.  This
effectiveness is leading to wider adoption in higher stakes domains,
which has the \emph{potential} to lead to more accurate, consistent,
and fairer decisions and the resulting societal benefits.  However,
given the higher stakes of these domains, there is a growing demand
that these systems provide explanations for their decisions, so that
necessary oversight can occur, and a citizen's due process rights are
respected~\cite{gdpr-goodman,gdpr-wachter,nyc,ainow-2017,kim-tutorial,finale,science-robotics-wachter,Caruana:2015,kushtst}.

The demand for explanation has manifested itself in new regulations that call for automated decision making systems to provide ``meaningful information'' on the logic used to reach conclusions \cite{gdpr-goodman,gdpr-wachter,SelbstP2017}. Selbst and Powles \shortcite{SelbstP2017}
interpret the concept of ``meaningful information'' as information that should be understandable to the audience (potentially individuals who lack specific expertise), is actionable, and  is flexible enough to support various technical approaches.

Unfortunately, the advance in effectiveness of machine-learning techniques has coincided with increased complexity in the inner workings of these techniques.  For some techniques, like deep neural networks or large random forests, even experts cannot explain how decisions are reached.  Thus, we have a stronger need for explainable AI, just when we have a greater gap in achieving it.

This has sparked a growing research community focused on this problem~\cite{whi-2017,whi-2018}.  Most of this research attempts to explain the inner workings of a machine learning model either directly, indirectly via a simpler proxy model, or by probing the model with related inputs. 
This paper proposes a different approach that requires a model to jointly produce both a decision as well as an explanation, rather than exposing the inner details of how the model produces a decision. The 
explanation is not constrained to any particular format and can vary to accommodate user needs.

\ignore{
However, inherent biases in the training data may still result in
biased decisions~\cite{propublica,NIPS2017_6988}, which, depending on the severity of imbalance in the dataset, may be
amplified beyond common human bias.  Because of this
concern, and the intuitive right of due process, there is a growing
need for such systems to provide explanations for their recommendations ~\cite{gdpr-goodman,gdpr-wachter,nyc,ainow-2017,kim-tutorial,finale,science-robotics-wachter,Caruana:2015,kushtst}.
}

The main contributions of this work are as follows:
\begin{itemize}
\item A description of the challenges in providing meaningful explanations for machine learning systems.

\item A new framework, called TED, that enables machine learning algorithms to provide meaningful explanations that match the complexity and domain of consumers. 

\item A simple instantiation of the framework that demonstrates the generality and simplicity of the approach.

\item Two illustrative examples and results that demonstrate the effectiveness of the instantiation in providing meaningful explanations.

\item A discussion on several possible extensions and open research opportunities that this framework enables.

\end{itemize}

The rest of this paper is organized as follows.
Section~\ref{sec-challenges} explores the challenges presented by the problem statement of providing explanations for AI decisions.
Section~\ref{sec-related} discusses related work.
Section~\ref{sec-ted} describes our general approach, TED, and a simple instantiation for providing explanations that are understandable by the consumer and discusses the advantages of the approach.
Section~\ref{sec-examples} presents results from two examples that demonstrate the effectiveness of the simple instantiation.
Section~\ref{sec-disc} discusses future directions and open issues for the TED approach.
Section~\ref{sec-conc} draws conclusions.

\section{Challenges to Providing AI Explanations}
\label{sec-challenges}


This section explores the challenges in providing meaningful explanations, which provide the motivation for the TED framework.

The concept of an explanation is probably as old as human
communication.  Intuitively, an explanation is the communication from
one person (A) to another (B) that provides justification for an
action or decision made by person A.  
%
%
\ignore{
This justification is usually
made informally, and is often iterative, with frequent checking by both
participants, using variants of {\em ``Do you understand?''} (A) or
{\em ``I don't understand''} (B).
}
Mathematicians use proofs to formally provide explanations.
These are constructed using agreed-upon logic and formalism, so that any
person trained in the field can verify if the proof/explanation is
valid.  Unfortunately, we do not have such formalism for non-mathematical
explanations. Even in the judicial system, we utilize
nonexpert jurors to determine if a defendant has violated a law, relying on their intuition and experience in weighing (informal) arguments made by prosecution and defense.

Since we do not have a satisfying formal definition for valid human-to-human explanations, developing one for system-to-human explanations is challenging~\cite{kim-tutorial,lipton2016mythos}.
\ignore{
This work focuses on providing explanations for a particular
recommendation, (called prediction or local explanation), which is
distinct from an explanation of the operation of an AI system
independent of any specific input (model or global
explanation)\cite{lime,causal-framework}.
}
%
%
\ignore{Explanations communicate a justification for a decision or action from
one entity to another that can be used to trust the decision.  The consumer of an explanation needs to be able to {\em
understand} the explanation, i.e., it needs to be meaningful to the consumer.  }
Motivated by the concept
of meaningful information \cite{gdpr-goodman,gdpr-wachter,SelbstP2017},
we feel that explanations must have the following three characteristics:

\begin{description}
\item [Justification:] An explanation needs to provide justification for a decision that increases trust in the decision.
This often includes some information that can be verified by the consumer.

\item [Complexity Match:] 
The {\em complexity of the explanation}
needs to match the complexity capability of the consumer~\cite{too-much,tip}.
For example, an explanation in equation form may be appropriate for a statistician, but not for a nontechnical person~\cite{miller2017inmates}.
\ignore{Furthermore, simple, representative explanations can be preferred to more complex,
but complete ones~\cite{miller2017explanation-review,miller2017inmates}.
}

\ignore{
Some US credit monitoring services utilize a FICO score to assess
credit risk.  FICO discloses key factors that impact the score, but
does not
provide the precise equation for
how these factor are weighted, nor the thresholds needed to achieve
the various credit levels
\cite{https://www.fico.com/en/resource-download-file/4292}
http://www.fico.com/en/products/fico-score-open-access/#news_releases
}

\item [Domain Match:] An explanation needs to be {\em tailored to the domain}, incorporating the relevant terms of the domain.  For example, an explanation for a medical diagnosis needs to use terms relevant to the physician (or patient) who will be consuming the prediction.
\ignore{Saying a treatment is recommended because ``$X1 > 0.45$'' and ``$X2 <
43.5$'' is not useful, no matter the mathematical
sophistication of the user, because $X1$ and $X2$ have no meaning to
the physician.  However, an explanation of ``glucose counts are
too high and blood pressure is too low'' is more informative.}

\end{description}

There are at least four distinct groups of people who are interested
in explanations for an AI system, with varying motivations.

\begin{description}
\item [Group 1: End User Decision Makers:]\label{decision-makers}
These are the people who use the recommendations of an AI system
to make a decision, such as physicians, loan officers, 
managers, judges, social workers, etc.
They desire explanations that can build their trust and confidence in
the system's recommendations and possibly provide them with additional
insight to improve their future decisions and understanding of the phenomenon.

\item [Group 2: Affected Users:]\label{affected-users}
These are the people impacted by the recommendations made by an AI
system, such as patients, loan applicants,
employees, arrested individuals, at-risk children, etc.
They desire explanations that can help them
understand if they were treated fairly and what factor(s)
could be changed to get a different result~\cite{finale}.

\item [Group 3: Regulatory Bodies:]\label{regulatory}
Government agencies, charged to
protect the rights of their citizens, want to ensure that decisions are made in a safe and fair manner, and that society is not negatively impacted by the decisions. 

\item [Group 4: AI System Builders:]\label{debuggers} Technical individuals (data scientists and developers) who build or deploy an AI system want to know
if their system is working as expected, how to diagnose and improve
it, and possibly gain insight from its decisions.
\end{description}

Understanding the motivations and expectations behind each group's needs for an
explanation will help to ensure a solution that satisfies these expectations.
For example, Group 4 is likely to desire a more complex
explanation of the system's inner workings to take action.
Group 3's needs may be satisfied by showing the overall process,
including training data, is fair and free of negative societal impact and they may not be able to consume the same level of complexity as Group 4. 
Group 1 will have a high need for domain sophistication, but will also have less tolerance for complex explanations.  Finally, Group 2 will have the lowest threshold for both complexity and domain information.  These are affected users, such as loan applicants, and need to have the reasons for their outcomes such as loan denials explained in a simple manner without industry terms or complex formulas.

\ignore{
Another complication is that this model lives
within a larger application that can further impact the decision, and
thus, its impact would need to be considered in any generated explanation.
}

In summary, outside of a logical proof, there is no clear definition of a valid explanation; it seems to be subjective to the consumer and circumstances.  Furthermore, there is a wide diversity of potential consumers of explanations, with different needs, different levels of sophistication, and different levels of domain knowledge.  This seems to make it impossible to produce a single meaningful explanation without any information
about the consumer.

\ignore{
Although ensuring fairness is often motivation for requesting explanations,
they are two distinct concepts.  Providing an explanation doesn't
guarantee a system will be fair and you can have fair systems that don't produce
explanations.
}

\ignore{
The impact from decisions can vary greatly~\cite{kush-data-science}.
Choosing a drug treatment, denying a promotion, or suggesting a
sentencing can have tremendous life consequences for the individuals
involved directly and indirectly.  In contrast, decisions regarding
what advertisement to show, what news story to recommend, or what
movie to watch next are usually not life-changing decisions for the
individuals involved.  Thus, there is a spectrum on the impact an
automated decision can have.  The higher the impact, the more likely
there is a need for explanations from Groups 1 and 2.  Group 4 will 
likely have a vested interest in any AI system that brings value to
society or is crucial to a business need.
}

\section{Related Work} 
\label{sec-related}
Prior work in providing explanations can be partitioned into three areas:

\begin{enumerate}
\item \label{interp}
Making existing or enhanced models {\em interpretable}, i.e.\ to provide a precise description of how the model determined its
decision (e.g.,~\cite{RibieroSG2016,montavon2017methods,unifiedPI}).

\item \label{2nd-model}
Creating a second, simpler-to-understand model, such as a small number of logical expressions, that mostly 
matches the decisions of the deployed model (e.g.,~\cite{bastani2017interpreting,Caruana:2015}).

\ignore{
\item \label{rationale}
Leveraging ``rationales'', ``explanations'', ``attributes'', or other ``privileged information'' in the training data 
to help improve the accuracy of the
algorithms (e.g.,~\cite{sun-dejong-2005,Zaidan07using-annotator,zaidan-eisner:2008:gen,DBLP:journals/corr/ZhangMW16,McDonnel16why-relevant,rationales,localizedattributes,peng-vision} 
}

\item \label{gen-rationale}
Work in the natural language processing and computer vision domains that generate rationales/explanations derived from input text (e.g., \cite{lei2016rationalizing,Yessenalina:2010:AGA:1858842.1858904,hendricks-2016}).

\end{enumerate}

The first two groups attempt to precisely describe how a
machine learning decision was made, which is particularly relevant for
AI system builders (Group 4).  The insight into the inner workings of a model can be used to improve the AI 
system and may serve as the seeds for an explanation to a non-AI expert.
However, work still remains to determine if these seeds are sufficient
to satisfy the needs of the diverse collection of non-AI experts (Groups 1--3).  
Furthermore, when the underlying features are not human comprehensible, these approaches are inadequate for providing human consumable explanations.

\ignore{
The third group, like this work, leverages additional information (explanations) in the training data, but with different goals.  The third group uses the explanations to create a more accurate model; we leverage the explanations to teach how to generate explanations for new predictions.  
}

The third group seeks to generate textual explanations with predictions. For text classification, this involves selecting the minimal necessary content from a text body that is sufficient to trigger the classification. For computer vision~\cite{hendricks-2016}, this involves utilizing textual captions in training to automatically generate new textual captions of images that are both descriptive as well as discriminative. 
\ignore{
While serving to enrich an understanding of the predictions, these systems do not necessarily facilitate an improved ability for a human user to understand system failures. } 
Although promising, it is not clear how these techniques generalize to other domains and if the explanations will be meaningful to the variety of explanation consumers described in Section~\ref{sec-challenges}.

Doshi-Velez et al.~\shortcite{finale} discuss the societal, moral, and legal expectations of AI explanations, provide
guidelines for the content of an explanation, and recommend that
explanations of AI systems be held to a similar standard as humans.  Our
approach is compatible with their view.   Biran and Cotton \shortcite{biran-cotton-2017} provide an excellent overview and taxonomy of explanations and justifications in machine learning.

Miller \shortcite{miller2017explanation-review} and Miller, Howe, and Sonenberg \shortcite{miller2017inmates}
argue that explainable AI solutions need to meet the needs of the
users, an area that has been well studied in philosophy, psychology,
and cognitive science.  They provides a brief survey of the most
relevant work in these fields to the area of explainable AI.
They, along with Doshi-Velez and Kim \shortcite{rsi}, call for more rigor
in this area.

\section{Teaching Explanations} \label{sec-ted}
\ignore{
The most straightforward approach to providing an explanation is the
one that group~\ref{debuggers} would take: provide a detailed description of how the system constructed
a recommendation.  This is quite similar to debugging a non-AI
application, where the application would be run with a debugger
enabled, similar to when a physician monitors a person's lungs with a
stethoscope, i.e., the system is monitored while it is functional to
look for ``interesting'' behavior. }

Given the challenges to developing meaningful explanations for the diversity of consumers described in Section~\ref{sec-challenges}, we 
advocate a non-traditional approach.  We suggest a high-level framework, with one simple instantiation, that we see
as a promising complementary approach to the traditional ``inside-out'' approach to providing explanations.  


\ignore{
Figure~\ref{fig-human-model} illustrates this point.

\begin{figure}
    \centering
    \includegraphics[width=3.2in]{humanmodel.png}
    \caption{Illustration of roles of Model, Application, and User}
    \label{fig-human-model}
\end{figure}
}

\ignore{
We approach the explanation problem differently.  Instead of 1) trying to
understand the actual details of an AI model and the system it lives
in and 2) trying to map this understanding to a complexity level and
domain a human can understand, we, instead, ask the explanation
consumer to train a system by demonstrating valid explanations in the
training data and use this training to generate explanations for the
system's recommendations.  This approach will increase the likelihood that explanations
provided by the system will match the complexity level and domain of
the explanation consumer.  
}

To understand the motivation for the TED approach, consider the common situation when a new employee is being trained for their new job, such as a loan approval officer.  The supervisor will show
the new employee several example situations, such as loan applications, and teach them the correct action: approve or reject, and explain the reason for the action, such as ``insufficient salary''.
Over time, the new employee will be able to make independent decisions on new loan applications and will give explanations based on the explanations 
they learned from their supervisor.
This is analogous to how the TED framework works.  We ask the training dataset to teach us, not only how to get to the correct answer (approve or reject), but also to provide the correct explanation, such as ``insufficient salary'', ``too much existing debt'', ``insufficient job stability'', ``incomplete application'', etc.  
From this training information, we will generate
a model that, for new input, will predict answers and provide explanations based on the explanations it was trained on.  Because 
these explanations are the ones that were provided by the training, they are
relevant to the target domain and meet the complexity level of the
explanation consumer.

Previous researchers have demonstrated that providing explanations with the training dataset may not add much of a burden to the training
time and may improve the overall accuracy of the training
data~\cite{Zaidan07using-annotator,zaidan-eisner:2008:gen,DBLP:journals/corr/ZhangMW16,McDonnel16why-relevant}.

\ignore{
To help address the matching {\em complexity} and {\em domain}
requirements described in Section~\ref{sec:requirements}, we 
augment the machine learning training process to ask the domain expert
to train the system on what is a valid explanation.
More specifically, we require each instance of the training data to
contain an explanation in addition to the class (decision).   We will
use these training explanations to provide explanations on new data
presented to the  model.
}

\subsection{\ted Framework and a Simple Instantiation}
The \ted
 framework leverages existing machine learning technology in a
straightforward way to generate a classifier that produces explanations
along with classifications.  To review, a supervised machine learning algorithm takes a training dataset that consists of a series of instances with the following two components:
\begin{description}
\item [X,] a set of features (feature vector) for the particular entity, such as an image, a paragraph, loan application, etc. 

\item [Y,] a  label/decision/classification for each feature
vector, such an image description, a paragraph summary,
or a loan-approval decision.
\end{description}

\noindent
The \ted framework requires a third component: 

\begin{description}
\item [E,] an explanation for each decision, $Y$, which can take any form, such as a number, text string, an image, a video file, etc.  Unlike traditional approaches, $E$ does not necessarily need to be expressed in terms of $X$.  It could be some other high-level concept specific to the domain that applies with some domain-specific combination of $X$, such as ``scary image'' or ``loan information is not trustworthy''.  Regardless of the format, we represent each unique value of $E$ with an identifier.
\end{description}

\begin{figure}
    \centering
    \includegraphics[width=3.25in]{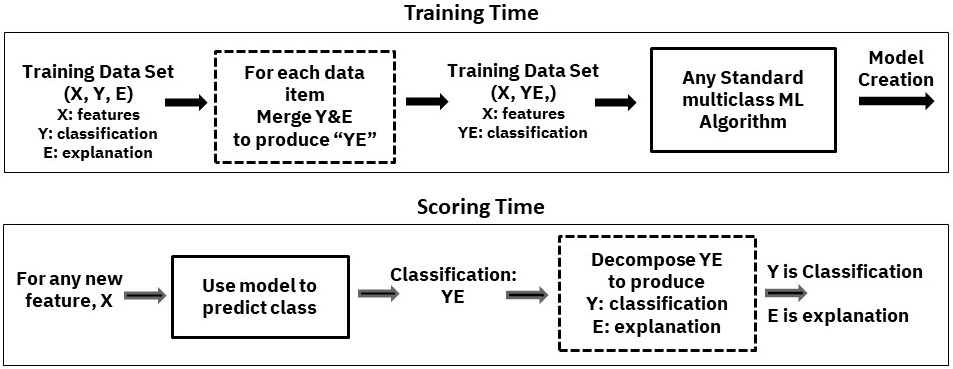}
    \caption{Overview of TED Algorithm \hfill {\tiny \copyright IBM Corporation}}
    \label{fig-ted-flowl}
\end{figure}

The \ted framework takes this augmented training set and produces a classifier that predicts both $Y$ and $E$.  There are several ways that this can be accomplished.  The instantiation we explore in this work is a simple Cartesian product approach.  This approach 
encodes $Y$ and $E$ into a new classification, called $YE$, which, along with the feature vector, $X$, is provided as the training input to any machine learning classification algorithm to produce a classifier that predicts $YE$'s.  After the model produced by the classification algorithm makes a prediction, we apply a decoding step to partition a $YE$ prediction into its components, $Y$ and $E$, to return to the consumer. Figure~\ref{fig-ted-flowl} illustrates the algorithm.  The boxes in dashed lines are new TED components that encode $Y$ and $E$ into $YE$ and decode a predicted $YE$ into its individual components, $Y$ and $E$.   The solid boxes represent 1) any machine learning algorithm that takes a normal training dataset: features and labels, and 2) the resulting model produced by this algorithm.

\ignore{
\begin{figure*}
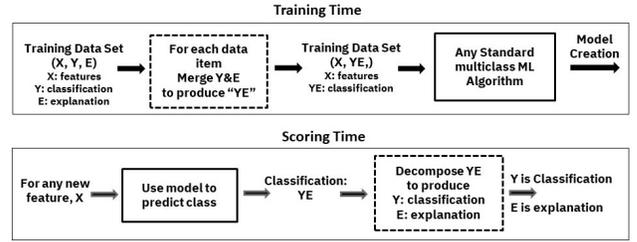
 
\begin{center}
\begin{small}
\begin{tabbing}
{\bf Input:} \=$X$: set of training feature vectors  \\
\> $Y$: set of correct classes for each feature vector \\
\> $Z$: set of explanations for each feature vector  \\

{\bf Output:} $X_{real}$, the real feature vectors to classify \\
	\>  $Y_{predict}$, the predicted classes for each feature
vector \\
  	\> $Z_{predict}$, the predicted explanations for each feature \\
	\> vector and class prediction \\
\hspace{2em} \\

{\bf Step 1:} Combine each instance of $Y$ and $Z$ into a unique $YZ$ \\
\hspace{2em} \\

{\bf Step 2:} Use $(X, YZ)$ as training input into any standard 
supervised ML classification algorithm (SVM, NN, etc.), \\
producing a classifier, C, where $X$ is the set of feature vectors 
 and $YZ$ is the correct class for those features \\
\hspace{2em} \\

{\bf Step 3:} Use C to classify new data instances, represented as
	feature vector, $X_{real}$, producing a class, $Y_{output}$ \\
\hspace{2em} \\

{\bf Step 4:} Decompose $Y_{output}$, using the inverse process of
step 1, 
to produce the predicted class  $Y_{prediction}$ and the\\
 predicted explanation $Z_{prediction}$.  
\end{tabbing}
\end{small}
\end{center}
\caption{Overview of \ted Algorithm}
\label{fig:alg}
\end{figure*} 
}

\subsection{Example}
Let's assume we are training a system to recommend cancer treatments.
A typical training set for such a system would be of the following form, where
$P_i$ is the feature vector representing patient $i$ and $T_j$,
represents various treatment recommendations.

\begin{small}
\begin{center}
\(
\begin{array}{c}
(P_1, T_A),
(P_2, T_A),
(P_3, T_A),
(P_4, T_A)\\
(P_5, T_B),
(P_6, T_B),
(P_7, T_B),
(P_8, T_C)
\end{array}
\)
\end{center}
\end{small}

The \ted approach would require adding an additional explanation component to the training dataset as follows: 

\begin{small}
\begin{center}
\(
\begin{array}{c}
(P_1, T_A, E_1),
(P_2, T_A, E_1),
(P_3, T_A, E_2),
(P_4, T_A, E_2)\\
(P_5, T_B, E_3),
(P_6, T_B, E_3),
(P_7, T_B, E_4),
(P_8, T_C, E_5)
\end{array}
\)
\end{center}
\end{small}

Each $E_i$ would be an explanation to justify why a feature vector
representing a patient would map to a particular treatment.  Some treatments could be recommended for multiple reasons/explanations.
For example, treatment $T_A$ is recommended for two different
reasons, $E_1$ and $E_2$, but treatment $T_C$ is only recommended for
reason $E_5$.

Given this augmented training data, the Cartesian product instantiation of the \ted framework
transforms this triple into a form that any supervised machine
learning algorithm can use, namely (feature, class) by combining the
second and third components into a unique new class as follows:

\begin{small}
\begin{center}
\(
\begin{array}{c}
(P_1, T_AE_{1}),
(P_2, T_AE_{1}),
(P_3, T_AE_{2}),
(P_4, T_AE_{2})\\
(P_5, T_BE_{3}),
(P_6, T_BE_{3}),
(P_7, T_BE_{4}),
(P_8, T_CE_{5})
\end{array}
\)
\end{center}
\end{small}


\begin{figure}
    \centering
        \includegraphics[width=3in]{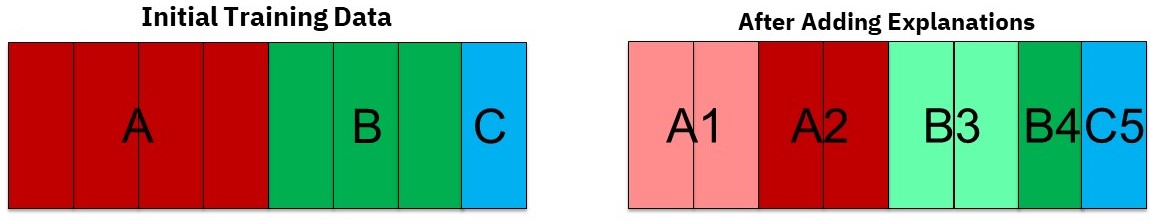}
    \caption{Changes to Training Dataset \hfill {\tiny \copyright IBM Corporation}}
    \label{fig-classpicture}
\end{figure}

Figure~\ref{fig-classpicture} shows how the training dataset would
change using the \ted approach for the above example.  The left picture
illustrates how the original 8 training instances in the example are mapped into the 3
classes.  The right picture shows how the training data is changed,
with explanations added.  Namely, Class $A$ was decomposed to
Classes $A1$ and $A2$.  Class $B$ was transformed into Classes $B3$
and $B4$ and Class $C$ became $C5$.

As Figure~\ref{fig-classpicture} illustrates, adding explanations
to training data implicitly creates a 2-level hierarchy in that the
transformed classes are members of the original classes, e.g.,
Classes $A1$ and $A2$ are a decomposition of the original Class $A$.
This hierarchical property could be exploited by employing hierarchical
classification algorithms when training to improve accuracy.

\subsection{Advantages} \label{sec-pro-con}
Although this approach is simple, there are several nonobvious advantages that are particularly important in addressing the requirements of explainable AI
for groups 1, 2, and 3 discussed in Section~\ref{sec-challenges}.

\noindent \textbf{Complexity/Domain Match:}
Explanations provided by the algorithm are guaranteed 
to match the complexity and mental model of the domain, given that
they are created by the domain expert who is training the system.\\
\noindent \textbf{Dealing with Incomprehensible Features:}
Since the explanation format can be of any type, they are not limited to
being a function of the input features, which is useful when the features are not comprehensible. \\
\noindent \textbf{Accuracy:}
Explanations will be accurate if the training data explanations are accurate and
representative of production data.\\
\noindent \textbf{Generality:}
This approach is independent of the machine learning
classification algorithm; it can work with any supervised
classification algorithm, including neural networks, making this
technique widely deployable.\\
\noindent \textbf{Preserves Intellectual Property:} \label{Ad-no-source-code}
There is no need to expose details of machine learning algorithm
to the consumer.  Thus, proprietary technology can remain protected by
their owners.\\
\noindent \textbf{Easy to incorporate:} The Cartesian product approach does not require a change to the current machine learning algorithm, just the addition of pre- and post-processing components: encoder and decoder.  Thus, an enterprise does not need to adopt a new machine learning algorithm, just to get explanations.\\
\noindent \textbf{Educates Consumer:}
The process of providing good training explanations will help properly set expectations for what kind of explanations
the system can realistically provide.  For example, it is probably easier to
explain in the training data why a particular loan application is
denied than  to explain why a particular photo is a cat.  
Setting customer expectations correctly for what AI systems can
(currently) do is important to their satisfaction with the system.\\
\noindent \textbf{Improved Auditability:}
After creating a \ted dataset, the domain expert will have enumerated all
possible explanations for a decision.  (The \ted system does not create
any new explanations.)  This enumeration can be useful for the
consumer's auditability, i.e., to answer questions such as ``What are the
reasons why you will deny a loan?''  or ``What are the situations in which you
will prescribe medical treatment X?''\\
\noindent \textbf{May Reduce Bias:}
Providing explanations will increase the likelihood of detecting
bias in the training data because 1) a biased decision will likely be
harder for the explanation producer to justify, and 2) one would expect that training instances with the same
explanations cluster close to each other in the feature space.
Anomalies from this property could signal a bias or a need for more
training data.


\section{Evaluation}
\label{sec-examples}

To evaluate the ideas presented in this work, we focus on two fundamental questions:
\begin{enumerate}
\item How useful are the explanations produced by the \ted approach? \label{ques-explanation}
\item How is the prediction accuracy impacted by incorporating explanations into the training dataset? \label{ques-accuracy}
\end{enumerate}

Since the \ted framework has many instantiations, can be incorporated into many kinds of learning algorithms, tested against many datasets, and used in many different situations, a definitive answer to these questions is beyond the scope of this paper.  Instead we try to address these two questions using the simple Cartesian product instantiation with two different machine learning algorithms (neural nets and random forest), on two use cases to show that there is justification for further study of this approach.  

Determining if any approach provides useful explanations is a challenge and no  
consensus metric has yet to emerge~\cite{finale}.  However, 
since the \ted approach requires explanations be provided for the target dataset (training and testing),   one can evaluate the accuracy of a model's explanation ($E$) in a similar way that one evaluates the accuracy of a predicted label ($Y$). 
\ignore{
In general, we expect several metrics of explanation efficacy to emerge, including those involving the target explanation consumers \cite{tip}.
}

The \ted approach requires a training set that contains explanations.  Since such datasets are not yet readily available, we evaluate the approach on two synthetic datasets described below: tic-tac-toe and loan repayment.

\subsection{Tic-Tac-Toe} \label{ttt}

\ignore{
We provide a synthetic data experiment using the tic-tac-toe dataset. This dataset contains the 4,520 legal non-terminal positions in this classic game.  Each position is labeled with a preferred next move ($Y$) and an explanation of the preferred move ($E$). Both $Y$ and $E$ were generated by a simple set of rules given in Section~\ref{ttt}. 
}

The tic-tac-toe example tries to predict the best move given a particular board configuration.
\ignore{
As an illustration of the proposed approach, we describe a simple domain, tic-tac-toe, where it is possible to automatically provide labels (the preferred move in a given board position) and explanations (the reason why the preferred move is best).  }
A tic-tac-toe board is represented by two $3 \times 3$ binary feature
planes, indicating the presence of X and O, respectively.  An
additional binary feature indicates the side to move, resulting in a
total of 19 binary input features.  Each legal non-terminal board position (4,520) is labeled with a preferred move, along with the reason the move is preferred.  The labeling is based on a simple set of rules that are executed  sequentially:\footnote{These rules do not guarantee optimal play.}
\begin{enumerate}
\item If a winning move is available, completing three in a row for the side to move, choose that move with reason \textit{Win}
\item If a blocking move is available, preventing the opponent from completing three in a row on their next turn, choose that move with reason \textit{Block}
\item If a threatening move is available, creating two in a row with an empty third square in the row, choose that move with reason \textit{Threat}
\item Otherwise, choose an empty square, preferring center over corners over middles, with reason \textit{Empty}
\end{enumerate}

Two versions of the dataset were created, one with only the preferred
move (represented as a $3 \times 3$ plane), the second with the
preferred move and explanation (represented as a $3 \times 3 \times 4$
stack of planes).  A simple neural network classifier was built on
each of these datasets, with one hidden layer of 200 units using ReLU
and a softmax over the 9 (or 36) outputs.  We use a 90\%/10\% split
of the legal non-terminal board positions for the training/testing datasets. 
This classifier obtained an accuracy of 96.5\% on the baseline move-only prediction task, i.e., when trained with just $X$ (the 19 features) and $Y$ it was highly accurate.

To answer the first question, does the approach provide useful explanations, we calculated the accuracy of the predicted explanation.  Although there are only 4 rules, each rule applies to 9 different preferred moves, resulting in 36 possible explanations.  Our classifier was able to generate the correct explanation 96.3\% of the time, i.e., very rarely did it get the correct move and not the correct rule.  

The second question asks how the accuracy of the classifier is impacted by the addition of $E$'s in the training dataset.  Given the increase in number of classes, one might expect the accuracy to decrease.  However, for this example, the accuracy of predicting the preferred move actually increases to 97.4\%.  This illustrates that the  approach works well in this domain; it is possible to provide accurate explanations without impacting the $Y$ prediction accuracy.  Table~\ref{table-results} summarizes the results for both examples.
 
\begin{table}
\begin{small}
\begin{center}
\caption{Accuracy for predicting Y and E in Tic-Tac-Toe and Loan Repayment}
\label{table-results}
\begin{tabular}{|c|c|c|c|c|}
\hline
& \multicolumn{4}{c|}{Accuracy (\%)} \\ \cline{2-5}
Training & \multicolumn{2}{c|}{Tic-Tac-Toe}
& \multicolumn{2}{c|}{Loan Repayment} \\ \cline{2-5}
Input	& Y &  E & Y  &  E  \\ \hline
X, Y & 96.5 & NA & 99.2 (0.2) & NA \\ \hline
X, Y, and E & 97.4 & 96.3 & 99.6 (0.1) & 99.4 (0.1) \\ \hline

\end{tabular}
\end{center}
\end{small}
\end{table}

\subsection{Loan Repayment}
The second example is closer to an industry use case and is based on the FICO Explainable Machine Learning Challenge dataset \cite{fico-challenge-2018}.  The dataset contains around 10,000 applications for Home Equity Line of Credit (HELOC), with the binary $Y$ label indicating payment performance (any 90-day or longer delinquent payments) over 2 years.  

Since the dataset does not come with explanations ($E$),\footnote{The challenge asks participants to provide explanations along with predictions, which will be judged by the organizers.} we generated them by training a rule set on the training data, resulting in the following two $3$-literal rules for the ``good'' class $Y=1$ (see \cite{fico-challenge-2018} for a data dictionary):
\begin{enumerate}
    \item NumSatisfactoryTrades $\ge$ 23 AND\\
ExternalRiskEstimate $\ge$ 70 AND\\ 
NetFractionRevolvingBurden $\le$ 63;
    \item NumSatisfactoryTrades $\le$ 22 AND\\
ExternalRiskEstimate $\ge$ 76 AND\\
NetFractionRevolvingBurden $\le$ 78.
\end{enumerate}
These two rules, from researchers at IBM Research, predict $Y$ with 72\% accuracy and were the winning
entry to the challenge~\cite{rules-neurips-2018}.  Since the TED approach
requires 100\% consistency between explanations and labels,  
we modified the $Y$ labels in instances where they disagree with the rules.
We then assigned the explanation $E$ to one of 8 values: 2 for the good class, corresponding to which of the two rules is satisfied (they are mutually exclusive), and 6 for delinquent, corresponding first to which of the rules should apply based on NumSatisfactoryTrades, and then to which of the remaining conditions (ExternalRiskEstimate, NetFractionRevolvingBurden, or both) are violated.

We trained a Random Forest classifier (100 trees, minimum 5 samples per leaf) on first the dataset with just $X$ and (modified) $Y$ and then on the enhanced dataset with $E$ added.  The accuracy of the baseline classifier (predicting binary label $Y$) was 99.2\%. The accuracy of TED in predicting explanations $E$ was 99.4\%, despite the larger class cardinality of 8.  In this example, $Y$ predictions can be derived from $E$ predictions through the mapping mentioned above, and doing so resulted in an improved $Y$ accuracy of 99.6\%.  While these accuracies may be artificially high due to the data generation method, they do show two things as in Section~\ref{ttt}: (1) To the extent that user explanations follow simple logic, very high explanation accuracy can be achieved; (2)
Accuracy in predicting $Y$ not only does not suffer but actually improves.  The second result has been observed by other researchers who have suggested adding ``rationales'' to improve classifier performance, but not for explainability~\cite{sun-dejong-2005,Zaidan07using-annotator,zaidan-eisner:2008:gen,DBLP:journals/corr/ZhangMW16,McDonnel16why-relevant,rationales,localizedattributes,peng-vision}.

\section{Extensions and Open Questions}
\label{sec-disc}
The \ted framework assumes a training dataset with explanations and uses it to train a classifier that can predict $Y$ and $E$.  This work described a simple way to do this, by taking the Cartesian product of $Y$ and $E$ and using any suitable machine learning algorithm to train a classifier. Another instantiation would be to bring together the labels and explanations in a multitask setting.  Yet another option is to learn feature embeddings using labels and explanation similarities in a joint and aligned way to permit neighbor-based explanation prediction.

Under the Cartesian product approach, adding explanations to a dataset increases the number of classes that the classification
algorithm will need to handle. 
This
could stress the algorithm's effectiveness or training time
performance, although we did not observe this in our two examples.  
However, techniques from the ``extreme classification'' community
~\cite{extreme-class-workshop} could be applicable.

Although the flexibility of allowing
any format for an explanation, provided the set of explanations can be enumerated, is quite general, it could encourage a large number
of explanations that differ in only unintended ways, such as
``insufficient salary'' vs.\ ``salary too low''.  Providing more structure via a domain-specific language (DSL) or good tooling could be useful.
If free text is used, we could leverage word embeddings to provide some
structure and to help reason about similar explanations. 

As there are many ways to explain the same phenomenon, it may be useful to explore having more than one version of the same base explanation for different levels of consumer sophistication.  Applications already do this for multilingual support, but in this case it would be multiple levels of sophistication in the same language for, say, a first time borrower vs.\ a loan officer or regulator.  This would be a postprocessing step once the explanation is predicted by the classifier.

Providing explanations for the full training
set is ideal, but may not be realistic.  Although it 
may be easy to add explanations \emph{while} creating the training dataset, it may be more challenging to add explanations after a dataset has been created because the creator may not available or may not remember the justification for a label.
One possibility is to use an external knowledge source to 
generate explanations, such as WebMD in a medical domain.  Another possibility is to request explanations on a subset of the training data and apply ideas from few-shot learning~\cite{Goodfellow-et-al-2016} to learn the rest of the training dataset explanations.  Another option is to use active learning to guide the user where to add explanations. One approach may be to first ask the
user to enumerate the classes and explanations and then to provide
training data ($X$) for each class/explanation until the algorithm
achieves appropriate confidence.  At a minimum one could investigate how the performance of the
explanatory system changes as more training explanations are
provided. Combinations of the above may be fruitful.

\section{Conclusions}
\label{sec-conc}
This paper introduces a new paradigm for providing explanations for machine learning model decisions.  Unlike existing methods, it does not attempt to probe the reasoning process of a model. Instead, it seeks to replicate the reasoning process of a human domain user.  
The two paradigms share the objective to produce a reasoned explanation, but the  model introspection approach is more suited to AI system builders who work with models directly, whereas the teaching explanations paradigm more directly addresses domain users. Indeed, the European Union GDPR guidelines say: ``The controller should find simple ways to tell the data subject about the rationale behind, or the criteria relied on in reaching the decision without necessarily always attempting a complex explanation of the algorithms used or disclosure of the full algorithm.'' 

Work in social and behavioral science~\cite{simplicity,miller2017inmates,miller2017explanation-review}
has found that people prefer explanations that are simpler,
more general, and coherent, even over more likely ones.
Miller writes that in the context of Explainable AI: ``Giving simpler
explanations that increase the likelihood that the observer 
both understands and accepts the explanation may be more useful to
establish trust~\cite{miller2017inmates}.''

Our two examples illustrate promise for this approach.  They both showed highly accurate explanations and no loss in prediction accuracy.
We hope this work will inspire other researchers to further enrich this paradigm.

\bibliographystyle{ACM-Reference-Format}
\bibliography{ExAbsent,ted,IEEEabrv}

\end{document}